\documentclass[conference]{IEEEtran}

\IEEEoverridecommandlockouts

\usepackage{cite}
\usepackage{amsmath,amssymb,amsfonts}
\usepackage{algorithmic}
\usepackage{graphicx}
\usepackage{textcomp}
\usepackage{xcolor}
\usepackage{booktabs}

\def\BibTeX{{\rm B\kern-.05em{\sc i\kern-.025em b}\kern-.08em
    T\kern-.1667em\lower.7ex\hbox{E}\kern-.125emX}}

\begin{document}
\title{Beyond Group Splits: Specimen-Level Cross-Validation and Visual Attribution
for Remaining-Shelf-Life Regression in Climacteric Fruit%
\thanks{This work is based on the research supported in part by the
National Research Foundation of South Africa (Ref Number
SRUG22051310444). The opinions, findings, conclusions or recommendations
are those of the author(s) alone, and the NRF accepts no liability
whatsoever in this regard.}}

\author{
\IEEEauthorblockN{Rovhona Mudau}
\IEEEauthorblockA{\textit{Department of Computer Science} \\
\textit{University of the Western Cape}\\
Cape Town, South Africa \\
4225313@myuwc.ac.za}
\and

\IEEEauthorblockN{Jean Frederic Isingizwe Nturambirwe}
\IEEEauthorblockA{\textit{eResearch Office} \\
\textit{University of the Western Cape}\\
Cape Town, South Africa \\
fisingizwe@uwc.ac.za}
\and

\IEEEauthorblockN{Clement Nthambazale Nyirenda}
\IEEEauthorblockA{\textit{Department of Computer Science /} \\
\textit{eResearch Office} \\
\textit{University of the Western Cape}\\
Cape Town, South Africa \\
cnyirenda@uwc.ac.za}
}

\maketitle
\begingroup
\renewcommand{\thefootnote}{}
\footnotetext{\textcopyright~2026 IEEE. Personal use of this material is permitted. Permission from IEEE must be obtained for all other uses, in any current or future media, including reprinting/republishing this material for advertising or promotional purposes, creating new collective works, for resale or redistribution to servers or lists, or reuse of any copyrighted component of this work in other works.}
\endgroup
\begin{abstract}

Estimating remaining shelf life (RSL) from images could provide
affordable decision support for perishable produce, but evaluation
protocols can substantially affect reported performance when repeated
images are available from the same biological specimen. We use the
Hass Avocado Ripening dataset, comprising 8,834 image--RSL pairs from
426 fruits across three storage regimes, to evaluate a frozen
ImageNet-pretrained visual backbone with a lightweight regression head.
Our contributions are threefold: we quantify the effect of
observation-level versus specimen-disjoint evaluation, compare
lightweight and heavier visual backbones under specimen-disjoint
cross-validation, and examine their spatial attributions using
Grad-CAM. Across ten observation-level random splits, the model
achieves a mean RMSE of 2.37 days with a standard deviation of
0.03 days, whereas specimen-disjoint 5-fold cross-validation yields
a mean RMSE of 3.12 days with a standard deviation of 0.11 days.
The corresponding mean coefficient of determination is 0.553.
A matched per-specimen comparison confirms higher error under
specimen-disjoint evaluation, with a probability value below 0.001
across 426 specimens, showing that observation-level partitioning gives a
substantially more optimistic estimate for this dataset and model
configuration. Under specimen-disjoint evaluation,
MobileNetV3-Small (0.93 million parameters) achieves accuracy
comparable to ResNet-18 while providing substantially higher
throughput, and Grad-CAM reveals differences in spatial attribution
between the lightweight backbones. These results support
specimen-disjoint evaluation and attribution analysis when assessing
lightweight vision models for longitudinal shelf-life prediction.
\end{abstract}


\section{Introduction}
\label{sec:intro}
Post-harvest loss is a significant issue in the food system, particularly for fresh fruits and vegetables that deteriorate quickly in poor storage conditions~\cite{fao2011}. A camera is the most cost-effective tool for predicting remaining shelf life (RSL), allowing operators to reroute, prioritise, or discount produce before deterioration occurs. The value of such a system, however, depends on a prior question: can its reported accuracy be trusted, and can a human verify that it is correct for the right reasons? The question is sharpened by the way these datasets are constructed. Monitoring deterioration requires photographing the same specimen repeatedly over time, so successive images of a single fruit are often highly similar. A standard random, image-level split can then place near-duplicate images of a test specimen into the training set, and the reported score reflects interpolation within a known distribution rather than generalisation to an unseen fruit. A recent peer-reviewed study trained convolutional networks on the same `Hass' avocado imagery, showing that the visual signal is learnable~\cite{xavier2024}; however, we take a different perspective by asking when such a predictor's performance is credible and when its logic is verifiable. The work is organised around three questions: how much reported accuracy survives when entire specimens are held out; whether a heavier backbone offers any advantage over a lightweight one under frozen-feature transfer; and whether the most accurate models attend to biologically meaningful fruit regions.

The rest of the paper is organized as follows: Section II reviews related work, Section III describes the materials and methods, Section IV presents the results, Section V discusses the findings, and Section VI concludes the paper.
\section{Related Work}
\label{sec:related}

Visual produce assessment is now dominated by learned convolutional
features~\cite{chuquimarca2024,mohanty2016}. Ripeness
\emph{classification} and shelf-life \emph{regression} are separated by
an important distinction: the former assigns a fruit to a discrete
class~\cite{begum2025,palakodati2020}, whereas the latter must resolve
fine-grained, time-varying cues into a calibrated estimate of remaining
time. The most direct precedent is Xavier et al.~\cite{xavier2024}, who
trained CNNs on the same longitudinal `Hass' avocado imagery and used
sample-disjoint training, validation, and test subsets for ripening-stage
prediction and subsequent shelf-life estimation. Other studies use environmental, physicochemical, or contextual
variables for shelf-life prediction~\cite{baskar2024,agarwal2025,
liang2025,mphahlele2026}. Mphahlele et al.~\cite{mphahlele2026},
for example, combined IoT-monitored environmental conditions with
laboratory quality attributes to evaluate machine-learning models for
fresh-apple shelf-life prediction. This paper investigates not only
whether the predictor attends to the fruit, but also how strongly its
reported RSL accuracy depends on the unit used to partition longitudinal
observations.

A major cause of poor reproducibility in applied machine learning is data
leakage~\cite{kapoor2023}. In agricultural vision, models can achieve high
in-distribution accuracy while relying on background or acquisition
artefacts rather than biological signal~\cite{noyan2022plantvillage,mohanty2016}.
Longitudinal shelf-life data introduce an additional repeated-measures
problem because several images may originate from the same biological
specimen. There are significant differences in the evaluation methods employed in related fruit studies. Hyperspectral sub-pictures from the same avocado photos were randomly assigned to training, validation, and test sets by Davur et al.~\cite{davur2023}. Without disclosing specimen-grouped partitioning for the regression analysis, Lee et al.~\cite{lee2025avocado} randomly divided 1,400 smartphone avocado images into 80/10/10 training, validation, and test partitions. In contrast, Xavier et al.~\cite{xavier2024} explicitly prevented observations from the same avocado sample from showing up in distinct subgroups for training, validation, and testing. Ahmad et al.~\cite{ahmad2026} supplemented internal cross-validation for mango shelf-life-stage classification with evaluation on an independent image dataset, while Brockelt et al.~\cite{brockelt2025} trained a strawberry storage-age model on control-storage samples and evaluated it on samples stored under modified-atmosphere packaging. Table~\ref{tab:evaluation_protocols} summarises these
differences.

\begin{table*}[t]
\centering
\caption{Evaluation protocols in representative fruit ripening and
shelf-life studies.}
\label{tab:evaluation_protocols}
\scriptsize
\setlength{\tabcolsep}{3pt}
\begin{tabular}{p{1.8cm} p{2.7cm} p{2.7cm} p{7.6cm}}
\hline
\textbf{Reference} &
\textbf{Task} &
\textbf{Protocol category} &
\textbf{Evaluation design} \\
\hline

\cite{davur2023} &
Hass avocado days-to-ripeness regression &
Random sub-image split &
Each hyperspectral fruit image's sub-images were assigned to training,
validation, and test sets; as a result, source-image identity was not
disjoint across partitions. \\

\cite{xavier2024} &
Hass avocado ripening classification and shelf-life estimation &
Specimen-level split &
To prevent observations from the same sample from crossing
partitions, avocado samples were uniquely assigned to 70/15/15 training,
validation, and test subsets. \\

\cite{lee2025avocado} &
Hass avocado firmness prediction with RSL recommendation &
Random image split &
A total of 1,400 images were randomly shuffled into 80/10/10 training,
validation, and test subsets; specimen grouping was not reported for the
regression split. \\

\cite{brockelt2025} &
Strawberry storage-age prediction &
Condition-level split &
Control-storage samples were used for model development and samples
stored under modified-atmosphere packaging were used as the test
condition. \\

\cite{ahmad2026} &
Mango shelf-life-stage classification &
Independent external validation &
Internal 10-fold cross-validation was supplemented by evaluation on a
separate online mango image dataset. \\

This study &
Hass avocado RSL regression &
Random, condition-level, and specimen-disjoint evaluation &
Observation-level random splitting is used as a leakage diagnostic;
storage regimes are held out for cross-condition analysis; primary
within-dataset generalisation is assessed using specimen-disjoint
5-fold cross-validation. \\
\hline
\end{tabular}
\end{table*}
The aforementioned protocols should not be used interchangeably because they address separate generalization issues. Related observations of the same biological specimen can occur in both training and test sets by means of random image-level partitioning. Instead, specimen-disjoint evaluation evaluates predictions on specimens that have not yet been seen under the specified acquisition and storage conditions. While independent external validation tests transfer to independently obtained data and hence address a wider distribution shift, condition-level holdout assesses transfer across storage regimes. Therefore, the controlled quantification of the performance optimism associated with observation-level specimen leakage and its comparison with specimen-disjoint and condition-level evaluation under an otherwise fixed prediction pipeline is the contribution of the current study rather than the introduction of specimen-disjoint partitioning itself, for which prior precedent exists.

There is also a persistent tension between predictive capacity and
deployment cost. Strong accuracy is achieved by residual and dense
backbones~\cite{he2016resnet,huang2018}, but heavy networks such as VGG
are unsuitable for edge hardware~\cite{simonyan2015}. The MobileNet
family is widely used on-device~\cite{boulahya2025,karim2024}, because
it trades a small accuracy margin for substantial efficiency gains
through depthwise-separable
convolutions~\cite{howard2017mobilenets,sandler2018mobilenetv2};
MobileNetV3 adds squeeze-and-excitation channel
attention~\cite{howard2019mobilenetv3}. Importantly, the
\emph{fine-tuning} regime~\cite{kornblith2019} provides much of the
evidence that ``better ImageNet models transfer better.'' Under
\emph{frozen} transfer, representations are reused without adaptation,
and a compact extractor may match a heavier one~\cite{yosinski2014}.
Lastly, explainability is becoming an important validation
requirement~\cite{arrieta2020}, and Grad-CAM is widely used for spatial
attribution~\cite{selvaraju2017gradcam,pal2025}. However, saliency maps
can be unfaithful~\cite{adebayo2018}, and squeeze-and-excitation blocks
reweight channels rather than spatial locations~\cite{hu2018senet},
motivating quantified attribution and perturbation-based corroboration
such as Score-CAM~\cite{wang2020scorecam}.

\section{Materials and Methods}
\label{sec:methods}

\subsection{Dataset and RSL Target Construction}
\label{sec:data}
We use the peer-reviewed `Hass' Avocado Ripening dataset~\cite{xavier2024}, which
tracks individual avocados under three storage regimes (cold storage at
$10\,^{\circ}\mathrm{C}$, denoted T10; T20; and ambient) with daily photographs and
a per-fruit ripening index. We construct a regression target using 14{,}710 photographs of 478 fruits. The first day a fruit reaches Ripening Index~4, corresponding to full spoilage, marks the end of its shelf life, and the number of days from acquisition to that day is the RSL of a photograph. We retain only images taken on or before the fruit's end-of-shelf-life day, and we exclude fruit that never reaches Index~4 because they are right-censored. This yields \textbf{8{,}834} image--RSL pairs from \textbf{426} fruit. All models are compared to a reference RMSE of \textbf{4.68}~days from a constant-mean predictor.

\subsection{Model and Training}
\label{sec:arch}
Each model uses a frozen ImageNet-pretrained convolutional backbone as a fixed
feature extractor, followed by a lightweight multilayer-perceptron (MLP) head
trained to regress RSL. We obtain features from the penultimate layer by replacing each backbone's classifier with an identity mapping, yielding 1280-, 576-, and 512-dimensional descriptors for MobileNetV2, MobileNetV3-Small, and ResNet-18, respectively. All backbones use \texttt{IMAGENET1K\_V1} weights and are frozen, so only the head is trained. The head is a two-layer MLP
($d\!\rightarrow\!128\!\rightarrow\!1$) with ReLU and dropout ($p=0.3$), optimised with Adam (learning rate $3\times10^{-4}$, weight decay $10^{-4}$) under mean-squared error for 150 epochs at batch size 128.

The pipeline is kept leak-free by two measures. First, only training-fold statistics are used to standardise features. Second, the regression target is standardised within each training fold to stabilise optimisation. Predictions are then inverse-transformed to days before scoring, ensuring that all reported errors are in the original unit. Using a unique random seed for each fold, the head is retrained separately for each fold.

\subsection{Evaluation Protocols}
\label{sec:protocols}
We compare three protocols that differ only in how images are partitioned:
\begin{enumerate}

\item \textbf{Random observation-level split:} without grouping by avocado identification, an 80/20 partition is performed at the image level. The procedure uses ten random seeds, and the regression head is retrained independently for each split.
Only the corresponding training partition is used to estimate the feature and target standardization parameters. The mean and standard deviation for each of the ten test partitions are used to report performance.
This approach is regarded as a leakage-prone diagnostic rather than an evaluation of performance on previously unseen specimens because images from the same avocado may appear in both training and test sets.

\item \textbf{Leave-one-storage-regime-out:}
one storage regime (T10, T20, or ambient) is held out entirely,
testing cross-condition generalisation.

\item \textbf{Specimen-disjoint 5-fold cross-validation:} the fruit identifiers are partitioned into five disjoint folds, each of
which is held out in turn. All observations associated with a given
avocado remain within the same fold, preventing specimen overlap
between training and test partitions and evaluating generalisation to
previously unseen specimens within the represented dataset and
acquisition conditions.

\end{enumerate}
For evaluation, only held-out observations are utilized. The regression head is retrained for every 80/20 split, and the mean and standard deviation of the ten held-out test sets are used to report performance. The random observation-level approach is repeated over ten independent seeds. Predictions from the three held-out storage conditions are combined over the corresponding test partitions for the leave-one-storage-regime-out evaluation. All observations from a specific avocado remain within a single fold for specimen-disjoint 5-fold cross-validation. Out-of-fold predictions for all 8,834 observations are obtained by combining predictions from the five held-out specimen folds.

The specimen-disjoint out-of-fold prediction for all images in a random test partition is selected for the paired comparison between random observation-level and specimen-disjoint assessment so that both protocols are assessed on identical image rows.
One paired error summary per avocado is produced for statistical testing by averaging the absolute errors inside each biological specimen and then over the ten random repetitions.

\subsection{Attribution and the Mass-on-Fruit Measure}
\label{sec:gradcam}
We calculate Grad-CAM~\cite{selvaraju2017gradcam} saliency for the two MobileNet backbones and summarise each heatmap by its \emph{mass-on-fruit}: the percentage of total saliency energy that falls inside the avocado silhouette, used to assess whether predictions rely on the fruit rather than the background. The silhouette is obtained by thresholding brightness and saturation against the light background. Because the fruit occupies a nearly constant area, about $17\%$ of the frame, a value near $0.17$ suggests chance-level attention. Values substantially above $0.17$ indicate spatial concentration on the fruit. For each of the four ripening stages, we compute mass-on-fruit for 40 images.

\subsection{Evaluation Metrics}
\label{sec:metrics}

We report image-level RMSE, MAE, and the coefficient of determination
$R^2$:

\begin{equation}
\begin{aligned}
\mathrm{RMSE}_{\mathrm{img}}
&=
\sqrt{
\frac{1}{N}
\sum_{i=1}^{N}
(\hat{y}_i-y_i)^2
},\\
\mathrm{MAE}_{\mathrm{img}}
&=
\frac{1}{N}
\sum_{i=1}^{N}
\left|\hat{y}_i-y_i\right|.
\end{aligned}
\label{eq:rmse_mae}
\end{equation}

\begin{equation}
R^2
=
1-
\frac{
\sum_{i=1}^{N}(y_i-\hat{y}_i)^2
}{
\sum_{i=1}^{N}(y_i-\bar{y})^2
},
\label{eq:r2}
\end{equation}

where $\hat{y}_i$ is the predicted RSL, $y_i$ is the true RSL,
$\bar{y}$ is the mean RSL, and $N$ is the number of held-out image
observations.

Because individual avocados contribute different numbers of
longitudinal images, pooled image-level metrics give greater weight to
specimens with longer observation sequences. We therefore also report
fruit-balanced RMSE and MAE. Let $S$ denote the number of held-out
biological specimens and let specimen $s$ contribute $n_s$
observations with residuals
$e_{si}=\hat{y}_{si}-y_{si}$. The fruit-balanced metrics are
\begin{equation}
\mathrm{RMSE}_{\mathrm{fruit}}
=
\sqrt{
\frac{1}{S}
\sum_{s=1}^{S}
\left(
\frac{1}{n_s}
\sum_{i=1}^{n_s}e_{si}^{2}
\right)},
\label{eq:fruit_rmse}
\end{equation}
\begin{equation}
\mathrm{MAE}_{\mathrm{fruit}}
=
\frac{1}{S}
\sum_{s=1}^{S}
\left(
\frac{1}{n_s}
\sum_{i=1}^{n_s}|e_{si}|
\right).
\label{eq:fruit_mae}
\end{equation}
Thus, all held-out observations are retained, but irrespective of how long an avocado's image sequence is, each avocado adds the same overall weight. Metrics are calculated independently for each held-out test set and provided as mean $\pm$ standard deviation across seeds for the ten random observation-level divisions.

As shown in Section~\ref{sec:results}, $R^2$ becomes unstable when a
held-out fold has little target variance. We therefore use RMSE and
MAE, particularly the fruit-balanced variants, as the more
interpretable measures for comparisons involving heterogeneous
storage conditions.

\section{Results}
\label{sec:results}

\subsection{Evaluation Protocol}

The reported performance was strongly affected by the evaluation
protocol (Table~\ref{tab:protocol}). Across ten independent
observation-level random splits, the model achieved
$\mathrm{RMSE}=2.37\pm0.03$\,d,
$\mathrm{MAE}=1.73\pm0.03$\,d, and
$R^2=0.744\pm0.006$. Under specimen-disjoint 5-fold
cross-validation, performance decreased to
$\mathrm{RMSE}=3.12\pm0.11$\,d,
$\mathrm{MAE}=2.35\pm0.08$\,d, and
$R^2=0.553\pm0.023$. For this dataset and model configuration, observation-level random partitioning produced a substantially more optimistic performance estimate, as demonstrated by the resulting difference in mean RMSE of approximately $0.76$\,d.

Fruits with longer observation sequences may be overweighted by pooled image-level metrics because the number of longitudinal observations varies among specimens. We therefore also report fruit-balanced RMSE
and MAE, in which errors are first aggregated within each avocado and
then averaged across the specimens represented in the corresponding
held-out evaluation set, so that each represented fruit contributes
equal total weight. Fruit-balanced RMSE/MAE under ten random observation-level splits were $2.42\pm0.04$/$1.76\pm0.02$\,d; specimen-disjoint 5-fold cross-validation produced $3.07$/$2.30$\,d; and leave-one-storage-regime-out evaluation produced $4.68$/$3.60$\,d. Therefore, the benefit of random observation-level partitioning cannot be explained by the fact that some specimens contribute more images than others; the protocol effect is still present even when the total weight of each fruit is identical.

Random-split and specimen-disjoint predictions were also assessed on matched held-out image observations to improve the statistical strength of the comparison. The corresponding specimen-disjoint out-of-fold prediction was selected for the same test rows for each random split. One paired error summary was produced for each biological specimen by averaging absolute errors within each avocado and across random-seed repetitions. Mean matched per-specimen MAE was
$1.77$\,d under random observation-level splitting and $2.32$\,d
under specimen-disjoint evaluation, corresponding to a mean paired
difference of $0.55$\,d. The difference was significant under a
Wilcoxon signed-rank test ($W=6828$, $p<0.001$, $n=426$ specimens).
The specimen-disjoint estimate was also stable across folds, with RMSE
ranging from $2.96$ to $3.27$\,d (standard deviation $0.11$\,d).
These results demonstrate that permitting repeated observations of the same fruit across training and test partitions can result in optimistic within-dataset performance estimates and that the greater specimen-disjoint error was not caused by a single held-out fold.

The leave-one-storage-regime-out result must be interpreted with care,
because strongly negative $R^2$ values can be misleading when the
held-out target variance is small. The three regimes yielded
RMSE/MAE values of $5.92/4.87$\,d (T10),
$3.82/2.85$\,d (T20), and $3.71/2.85$\,d (ambient).
The fast-ripening T20 and ambient regimes had mean RSL values of
approximately $3$\,d and relatively limited target variance; therefore,
moderate absolute errors produced strongly negative $R^2$ values
($-2.34$ and $-2.28$, respectively). In contrast, T10 had a mean RSL
of $7.75$\,d and produced the largest absolute errors when held out.
Because $R^2$ depends on the variance of the target within each held-out
condition, MAE is treated as the more interpretable statistic for
cross-condition comparison.

\begin{table*}[t]
\centering
\caption{Image-level and fruit-balanced performance under the three evaluation protocols.}
\label{tab:protocol}
\footnotesize
\setlength{\tabcolsep}{6pt}
\renewcommand{\arraystretch}{1.1}
\begin{tabular}{@{}lccccc@{}}
\toprule
\textbf{Protocol} &
\textbf{RMSE$_{\mathrm{img}}$ (d)} &
\textbf{MAE$_{\mathrm{img}}$ (d)} &
\textbf{$R^2_{\mathrm{img}}$} &
\textbf{RMSE$_{\mathrm{fruit}}$ (d)} &
\textbf{MAE$_{\mathrm{fruit}}$ (d)} \\
\midrule

Random observation-level split
& $2.37\pm0.03$
& $1.73\pm0.03$
& $0.744\pm0.006$
& $2.42\pm0.04$
& $1.76\pm0.02$ \\

Leave-one-storage-regime-out
& 5.18
& 4.07
& -0.227
& 4.68
& 3.60 \\

Specimen-disjoint 5-fold CV
& $3.12\pm0.11$
& $2.35\pm0.08$
& $0.553\pm0.023$
& 3.07
& 2.30 \\

\bottomrule
\end{tabular}
\end{table*}

\subsection{Backbone Comparison}
Under identical specimen-disjoint 5-fold evaluation, backbone capacity does not translate into better accuracy (Table~\ref{tab:rq2-backbones}). The smallest and fastest extractor, MobileNetV3-Small, with only $0.93$\, M parameters and a throughput of $76$ images/s, is at least as accurate as the larger alternatives ($R^2=0.596\pm0.027$); the differences among the three backbones are small relative to the observed fold-to-fold variability. The decisive point is therefore not that the lightest model wins but that the heavier ResNet-18 does not provide a corresponding improvement in predictive performance under the evaluated frozen-feature setting. This is consistent with the frozen-feature, modest-data regime: without fine-tuning, the larger network cannot exploit its extra capacity, whereas the efficient MobileNet representations are well matched to the fine-grained skin-colour signals that indicate ripening. Under the resource-constrained transfer setting targeted here, the lightweight architecture is therefore the rational default on both accuracy and compute, confirming that the fine-tuning evidence for heavier backbones~\cite{kornblith2019} need not apply to frozen transfer~\cite{yosinski2014}.

\begin{table}[tbp]
\centering
\caption{BACKBONE COMPARISON UNDER SPECIMEN-DISJOINT 5-FOLD CROSS-VALIDATION }
\label{tab:rq2-backbones}
\scriptsize
\setlength{\tabcolsep}{4pt}
\begin{tabular}{lccccc}
\toprule
\textbf{Backbone} & \textbf{Par.\,(M)} & \textbf{Thr.} & \textbf{RMSE (d)} & \textbf{MAE (d)} & \textbf{$R^2$} \\
\midrule
\textbf{MobileNetV3-S} & \textbf{0.93} & \textbf{76/s} & $\mathbf{2.97\!\pm\!0.14}$ & $\mathbf{2.24\!\pm\!0.12}$ & $\mathbf{0.596\!\pm\!0.027}$ \\
MobileNetV2 & 2.22 & --- & $3.12\!\pm\!0.11$ & $2.35\!\pm\!0.08$ & $0.553\!\pm\!0.023$ \\
ResNet-18   & 11.18 & 11/s & $3.09\!\pm\!0.08$ & $2.35\!\pm\!0.07$ & $0.562\!\pm\!0.018$ \\
\bottomrule
\end{tabular}
\end{table}

\subsection{Visual Attribution}
\begin{figure}[tbp]
\centering
\includegraphics[width=\columnwidth]{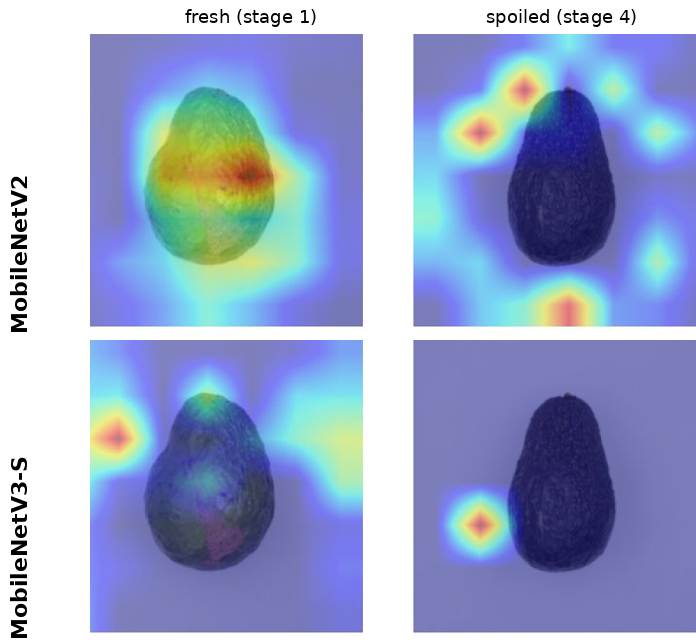}
\caption{Grad-CAM saliency at the ripening endpoints. MobileNetV2 (top) concentrates
on the fruit when fresh and disperses once spoiled; the more accurate
MobileNetV3-Small (bottom) localises to the background at both stages.}
\label{fig:gradcam}
\end{figure}

Accurate prediction does not ensure that a model attends to the fruit.
Table~\ref{tab:rq3-massonfruit} shows mass-on-fruit by ripening stage, where the chance baseline is $0.17$. When ripening cues are clear, MobileNetV2 focuses heavily on the fruit---mass-on-fruit reaches $0.44$ at stage~1, $2.6\times$ chance (Fig.~\ref{fig:gradcam})---and decreases monotonically to $0.12$ at full spoilage. This decrease makes physiological sense: saliency disperses because a spoiled, uniformly dark avocado lacks distinguishing surface characteristics. Despite having the most accurate predictions of the three backbones, MobileNetV3-Small behaves very differently. Its mass-on-fruit remains close to chance at every stage ($0.12$--$0.28$), with substantial variance and no discernible trend, localising to the background.

This reveals a divergence between the two lightweight models' accuracy and interpretability: MobileNetV3-Small, which is more accurate, is less spatially verifiable. We interpret this cautiously. The disparity is likely caused by MobileNetV3's squeeze-and-excitation blocks, which reweight feature channels globally rather than spatially~\cite{hu2018senet}. As a result, predictive evidence may be dispersed across channels in a way that gradient-based spatial attribution is unable to localise. Because saliency maps are not guaranteed to be faithful~\cite{adebayo2018}, the diffuse maps may partially reflect a limitation of Grad-CAM \emph{as computed on} squeeze-and-excitation architectures rather than a property of the model itself. Corroboration with perturbation-based attribution, such as Score-CAM~\cite{wang2020scorecam}, is left to future work. Despite its slightly lower accuracy, MobileNetV2's more localised Grad-CAM attribution may be advantageous
for inspection in decision-support settings where forecasts must be
checked.

\begin{table}[tbp]
\centering
\caption{Grad-CAM mass-on-fruit by stage}
\label{tab:rq3-massonfruit}
\footnotesize
\begin{tabular}{lcc}
\toprule
\textbf{Stage} & \textbf{MobileNetV2} & \textbf{MobileNetV3-Small} \\
\midrule
1 (fresh)    & \textbf{0.44} & 0.12 \\
2            & 0.29 & 0.23 \\
3            & 0.16 & 0.28 \\
4 (spoiled)  & 0.12 & 0.14 \\
\bottomrule
\end{tabular}
\end{table}

\section{Discussion}
\label{sec:discussion}

Three implications follow. First, the evaluation protocol has a larger
effect on reported performance than backbone choice in this
repeated-measures dataset. Across ten observation-level random splits,
RMSE was $2.37\pm0.03$\,d, compared with
$3.12\pm0.11$\,d under specimen-disjoint 5-fold
cross-validation, a difference of approximately $0.76$\,d. This
difference is larger than that observed between any two backbones in
Table~\ref{tab:rq2-backbones}. The matched analysis supports the same
conclusion: mean per-specimen MAE increased from $1.77$\,d under
random observation-level splitting to $2.32$\,d under
specimen-disjoint evaluation, with a mean paired difference of
$0.55$\,d ($p<0.001$, $n=426$ specimens). Therefore, neither a single favorable random partition nor a single highly difficult specimen fold can be attributed to the effect. Rather, allowing observations from the same avocado to occur in both training and test segments yields an optimistic estimate of performance for this dataset and model configuration. The observed difference should not be interpreted as a universal correction applicable to other shelf-life datasets.

The dependence among repeated images also affects metric aggregation.
Specimens that provide more observations are given higher weight in image-level evaluation. The same qualitative pattern persists when each avocado is instead given an identical total weight: the corresponding MAEs are $1.76$ and $2.30$\,d, and the fruit-balanced RMSE is $2.42$\,d for random observation-level assessment and $3.07$\,d for specimen-disjoint evaluation. Therefore, varying numbers of images per specimen are not the only explanation for the protocol effect. Because the held-out regimes have significantly different target distributions, interpretation of $R^2$ for leave-one-storage-regime-out evaluation requires extra caution. Specifically, minor absolute errors result in highly negative $R^2$ values due to the limited RSL variance in the T20 and ambient folds. For comparisons across heterogeneous storage regimes, MAE and RMSE are consequently easier to understand than $R^2$.

Second, while the feature extractor is frozen, the backbone comparison indicates limited benefit from greater model capacity. Despite having far fewer parameters and higher observed throughput, MobileNetV3-Small achieves an RMSE of $2.97\pm0.14$\,d during specimen-disjoint evaluation, compared with $3.09\pm0.08$\,d for ResNet-18.
Therefore, increasing backbone size does not result in a proportional gain in RSL prediction accuracy within the assessed transfer-learning setting. This result is relevant to resource-constrained deployment, where predictive performance may be impacted by inference cost, memory footprint, and throughput. It does not mean that a different adaptation method, such as task-specific fine-tuning, may not be advantageous for a larger backbone.

Third, predictive accuracy and spatial attribution quality do not
necessarily coincide. Among the assessed backbones, MobileNetV3-Small has the best prediction performance; however, its Grad-CAM maps are less reliably localized to the avocado than those generated by MobileNetV2. It is challenging to ascertain whether this discrepancy stems from the underlying model representation, the attribution method, or both, due to architectural mechanisms such as squeeze-and-excitation channel reweighting~\cite{hu2018senet} and the known limitations of gradient-based saliency methods~\cite{adebayo2018}. Therefore, rather than serving as evidence of the network's causal properties, the Grad-CAM results should be viewed as an attribution audit. Nevertheless, spatially plausible attributions may help identify obvious failure modes in
decision-support settings. These findings motivate considering
predictive performance, computational efficiency, and attribution
behaviour jointly when selecting lightweight models.

\section{Conclusion}
\label{sec:conclusion}

This study makes three main contributions to image-based longitudinal
remaining-shelf-life prediction.
Firstly, it quantifies the effect of an evaluation protocol within a defined modeling pipeline. RMSE was $2.37\pm0.03$\,d over ten observation-level random splits, whereas specimen-disjoint 5-fold cross-validation yielded $3.12\pm0.11$\,d. The same direction of effect was confirmed by a corresponding per-specimen study, which demonstrated that for this dataset and model configuration, observation-level random partitioning yields a substantially more optimistic performance estimate. Furthermore, the conclusion holds valid under fruit-balanced aggregate, suggesting that varying numbers of observations per specimen are not the sole explanation.

Second, the backbone comparison shows that greater model capacity does
not necessarily improve predictive performance under frozen-feature
transfer. With $0.93$\, M parameters, MobileNetV3-Small produced substantially greater throughput while achieving accuracy comparable to ResNet-18. When computational efficiency is crucial and more backbone capacity does not result in improved RSL prediction, this encourages the use of lightweight visual extractors.

Thirdly, the Grad-CAM analysis uncovers variations in spatial attribution behavior that predictive metrics alone fail to capture. MobileNetV3-Small demonstrated the highest predictive efficacy across the assessed architectures, but MobileNetV2 yielded more consistently localized Grad-CAM visualizations. These results are interpreted as an attribution audit rather than evidence that one backbone is inherently more faithful than another because saliency methods are unable to demonstrate causal feature usage on their own.

Several limitations bound these findings. The backbones were used as
frozen feature extractors, so full or partial fine-tuning could alter
their relative performance. The mass-on-fruit measure depends on an
estimated fruit silhouette obtained using a heuristic threshold, and
RSL is defined relative to a single end-of-shelf-life stage within one
produce type and acquisition setting.  The present results therefore
characterise within-dataset evaluation behaviour rather than external
generalisation. Future work should incorporate environmental sensor
streams associated with post-harvest deterioration, evaluate the
approach under independent produce and imaging conditions, and
corroborate the attribution findings using complementary
perturbation-based methods such as Score-CAM~\cite{wang2020scorecam}.

\bibliographystyle{IEEEtran}
\bibliography{references}

\end{document}